# Explainable Predictive Condition-based Maintenance of Naval Propulsion Systems using Fuzzy Logic

Dionisis Kalogeropoulos

Department of Computer Science and Biomedical Informatics, University of Thessaly, dkalogerop@uth.gr

Georgia Sovatzidi

Department of Computer Science and Biomedical Informatics, University of Thessaly, gsovatzidi@uth.gr

Panagiotis G. Kalozoumis

Department of Computer Science and Biomedical Informatics, University of Thessaly, pkalozoumis@uth.gr

Dimitris K. Iakovidis*

Department of Computer Science and Biomedical Informatics, University of Thessaly, diakovidis@uth.gr

The shipping industry has a significant impact on the global economy, emphasizing the need for operational availability and safety through the use of effective maintenance techniques. During the last decades, predictive maintenance (PdM) has emerged as a promising solution compared to the existing conventional maintenance systems. This is because it offers several advantageous functions, such as damage predictions for vessel components, reduced downtime, improved and extended life of machinery, as well as higher safety during voyages. However, existing methodologies developed for performing PdM do not provide explanations of their results to users, so that they can understand the failures that may occur. To address this limitation, this paper proposes a novel framework based on a fuzzy decision tree and a deep residual neural network, aiming to perform explainable PdM on naval vessels. The proposed framework is able to generate fuzzy local rules based on the dataset used, and can provide explanations of its outcomes, using cause-and-effect relationships, in a way that are understandable to users, thereby gaining their trust. Experiments using a publicly available dataset demonstrate the effectiveness of the proposed framework, as it achieves an accuracy of 99.24%.

## 1 INTRODUCTION

The operational efficiency, reliability and longevity of marine vessels, especially those equipped with complex propulsion systems, depend significantly on the effectiveness of predictive maintenance (PdM) strategies. PdM uses advanced computational intelligence techniques aiming to analyze real-time data and identify patterns and anomalies that could be early indicators of potential system failures. A key element of PdM is Condition-based Maintenance (CBM), which allows the planning of the corresponding maintenance actions based on the actual condition of the equipment rather than the defined schedules [1]. Furthermore, it is particularly critical in marine operations, as it continuously monitors and evaluates system health indicators, allowing maintenance to be performed only when required, thus reducing unnecessary downtime and operating costs. In this context, the health status of a naval propulsion system (NPS) is represented through degradation states, which describe progressive levels of performance deterioration with respect to normal operating conditions. These states capture the gradual evolution from healthy operation to degraded behavior caused by aging and wear mechanisms and indirectly inferred from available sensor measurements [2], [3].

* Corresponding author.

In recent years, only a limited number of studies have addressed the challenge of PdM in naval vessel propulsion systems. Specifically, *Cipollini et al*. presented a framework that utilized both supervised and unsupervised data analysis techniques to predict the performance decay of NPSs for CBM purposes, estimating the decay of the gas turbine, gas turbine compressor, hull, and propeller of the vessel [4]. *Gao et al*. proposed a deep learning framework, named DRN-GAN, which integrated a Deep Residual Network (DRN) and a Generative Adversarial Network (GAN) to address data imbalance,perform health degradation assessment and performance prediction for naval gas turbines by learning complex nonlinear relationships [5]. Moreover, *Javadnejad et al.* proposed a hybrid probabilistic framework, named BiGMM-HMM that combined a Bivariate Gaussian Mixture Model (BiGMM) with a Hidden Markov Model (HMM) to improve PdM performance in naval vessel propulsion equipment, while preventing overfitting [6]. *Sahraoui et al.* combined conventional machine learning approaches with ensemble learning strategies to perform fault diagnosis in gas turbine propulsion systems, and improve the diagnostic accuracy and generalizability of CBM for NPSs [7].

In this paper, we present a framework for predicting engine degradation in NPSs. The proposed framework is a teacher-guided model, consisting of a neural network (NN) with a fuzzy-logic-based classifier head. In the proposed framework, the teacher model is an NN that is trained on the data used and produces soft labels to characterize the degradation state of a NPS. A student model, which in this study is a Fuzzy Decision Tree (FDT) is trained on soft labels over instance-specific neighborhoods, aiming to produce explanations for the outcomes, in a way that is understandable by humans. Furthermore, the proposed framework includes a gradient-based explainability module that generates saliency maps to identify sensor inputs that lead to a degradation prediction. This work advances existing PdM methodologies by enabling inherently interpretable predictions through the combination of model distillation and rule-based explanations. This design enables transparent and human-readable insights into the decision-making process without compromising predictive performance.

The rest of the paper is organized as follows: Section 2 describes the methodology, whereas Section 3 presents the experiments, results, and two explainable examples of failure prediction in NPS. Finally, the conclusions derived from this study are summarized in Section 4.

## 2 METHODOLOGY

The proposed framework involves two phases: a training and a testing phase. In the training phase, an NN is trained to predict the degradation states of a NPS; in the testing phase, predictions are made and explained for specific input values, based on the corresponding defined fuzzy rules and generated saliency maps.

### 2.1 Training Phase

The proposed framework employs a deep NN classifier, which is the teacher model, comprising of a feature extractor with an input layer and three hidden layers, followed by a classification head. The feature extractor (Figure 1) consists of the input layer, which is implemented through a custom module, termed Dense Block, which is composed of a linear transformation followed by batch normalization and a dropout layer to provide regularization. The subsequent feature extractor layers utilize an extension of the Dense Block module, termed Residual Block, by incorporating additive skip connections that add the layer inputs to the corresponding outputs. This residual formulation facilitates improved gradient flow, thereby enhancing training stability and optimization performance [8]. The Fuzzy Classifier component integrates fuzzy logic into the prediction process through two consecutive layers; the Fuzzification layer and the Defuzzification layer that produces the final classification output. The proposed component implements a differentiable fuzzy membership function within a NN framework, following the paradigm of neuro-fuzzy systems, such as ANFIS [9] and its deep extensions [10], [11]. Each output neuron represents a fuzzy rule whose activation is computed as an exponential decay based on a transformed Euclidean distance ($\ell_2$-norm) [9]. For a given input vector, the layer applies a learnable affine transformation composed of scaling, rotation, and translation parameters, followed by an $\ell_2$-norm and exponential mapping to obtain membership degrees. This allows end-to-end training while retaining the uncertainty modeling of a fuzzy logic-based architecture. More specifically, the layer implements a generalized membership function [11]:

$$\mu_{c_i}(\boldsymbol{x}) = \exp\left(-|\boldsymbol{A}_{c_i}\boldsymbol{x} + \boldsymbol{b}_{c_i}|_2\right) \quad (1)$$

where $c_i, i = 1, \dots F$ and $F$ is the total number of fuzzy rules, $\mu_{c_i}(\boldsymbol{x})$ represents the membership degree of the input vector $\boldsymbol{x} \in \mathbb{R}^d$ to a $c_i$. The matrix $\boldsymbol{A}_{c_i} \in \mathbb{R}^{h\times h}$ includes the membership functions and is a learnable linear transformation that controls their orientation and scaling, while $\boldsymbol{b}_{c_i} \in \mathbb{R}^h$ defines the centroid of the fuzzy rule. The $\ell_2$-norm of the transformed input is mapped via the exponential function to a value in [0,1] such that smaller distances correspond to higher membership degrees.

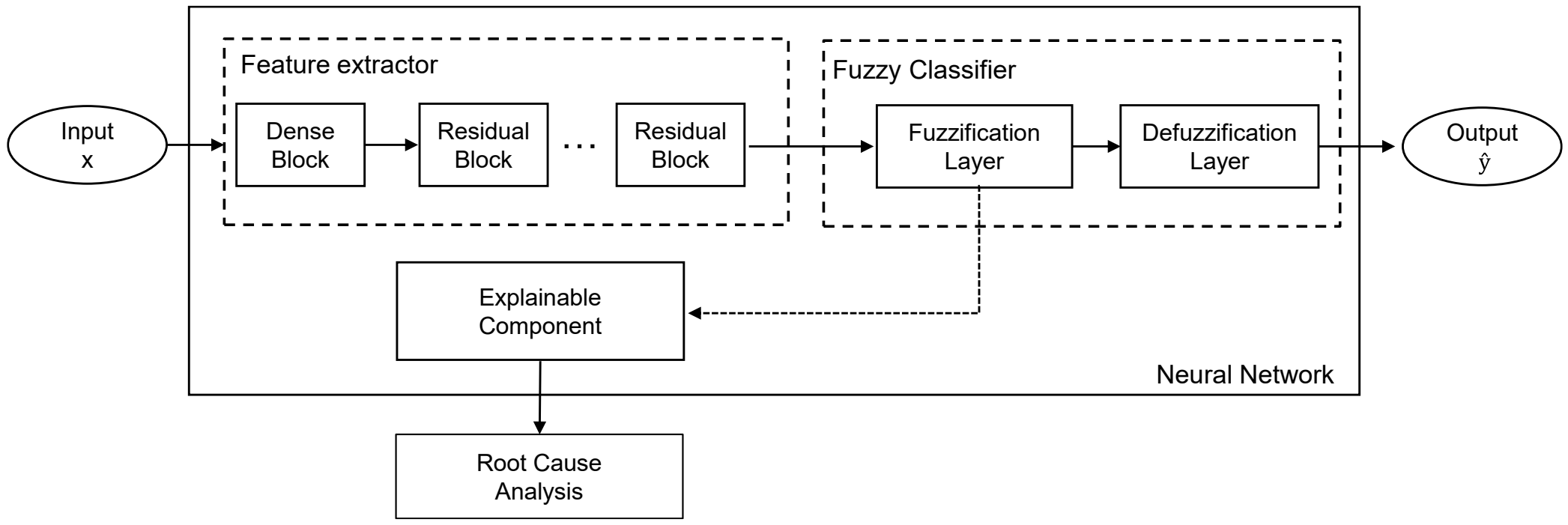

Figure 1: Teacher model architecture.

Additionally, the defuzzification process is also implemented through a learnable linear layer that aggregates the activations of all fuzzy rules to produce a final crisp output. Specifically, for an input $\boldsymbol{x}$, the defuzzified output $\widehat{\boldsymbol{y}}_{\boldsymbol{k}} \in \mathbb{R}^{d_{\text{out}}}$ is computed as a normalized weighted sum of the subsequent vectors of the rule $\boldsymbol{z}_{c_i} \in \mathbb{R}^{d_{\text{out}}}$[12]:

$$\widehat{\boldsymbol{y}}_{\boldsymbol{k}} = \frac{\sum_{i=1}^{F} \alpha_{c_i}(\boldsymbol{x})\cdot \boldsymbol{z}_{c_i}}{\sum_{i=1}^{F} \alpha_{c_i}(\boldsymbol{x})} \quad (2)$$

where $\alpha_{c_i}(\boldsymbol{x})$ denotes the activation of fuzzy rule $c_i$ for input $\boldsymbol{x}$. Furthermore, the proposed framework incorporates an explainable component that leverages the trained membership functions within the Fuzzification layer. For a given input $\boldsymbol{x}$, the gradient-based saliency is defined as the magnitude of the gradient of the output of a trained model $T$ with respect to the input features. Formally, let $p_T(\widehat{\boldsymbol{y}}_{\boldsymbol{l}} \mid \boldsymbol{x}), l = 1, \dots, L$ and $L$ corresponds to the total number of classes included in the dataset, denote the model output associated with the target class $\widehat{\boldsymbol{y}}_{\boldsymbol{l}}$. The saliency of the $i$-th input feature $x_i$ is defined as follows [13]:

$$S_i(\boldsymbol{x}) = \left|\frac{\partial p_T(\widehat{\boldsymbol{y}}_l|\boldsymbol{x})}{\partial x_i}\right|, i = 1, \dots, d \ (3)$$

The feature relevance is calculated from the gradients of dominant fuzzy rule activations (Eq. 4). Thus, to obtain localized, rule-based explanations, the proposed framework further conditions this analysis on the most relevant fuzzy rules, which are ranked based on their activation strength. Let $f(\boldsymbol{x})$ denote the latent feature representation produced by the preceding network layers and used as input to the fuzzification layer; $F_k, k \in \mathbb{N}$, denote the set of top-$k$ fuzzy rules with the highest membership values $\mu_c(\boldsymbol{x})$. The final feature relevance scores are computed by aggregating the absolute gradients of the corresponding rule activations with respect to the *i*-th input feature as expressed in Eq. (4) [12]:

$$\widetilde{S}_i(\boldsymbol{x}) = \sum_{c \in F_k} \left|\frac{\partial \mu_c(f(x))}{\partial x_i}\right|, i = 1, \dots, d \ (4)$$

By calculating the gradients of fuzzy rule activations with respect to the input, rule-specific saliency maps are generated and can provide interpretable feature importance maps, based on both fuzzy rule activations and gradient information.

### 2.2 Testing phase

During the testing phase (Figure 2), an input sample represented by a feature vector $\boldsymbol{x} \in \mathbb{R}^d$, where $d$ denotes the dimensionality of the input space, is processed by the teacher model (Figure 1). Saliency maps are then generated and are used to perform the root cause analysis. In addition, the input data is utilized to perform *k*-NN sampling, and its local neighborhood $\boldsymbol{N_k}$ is obtained. This local neighborhood and its corresponding teacher predictions $\{\boldsymbol{N_k}, \widehat{\boldsymbol{y}}_{\boldsymbol{k}}\}$ form the training data for the FDT used. The generated samples constitute the neighborhood nearest to the input $\boldsymbol{x}$ and, based on the DFT, the most informative local fuzzy IF-THEN rules are defined.

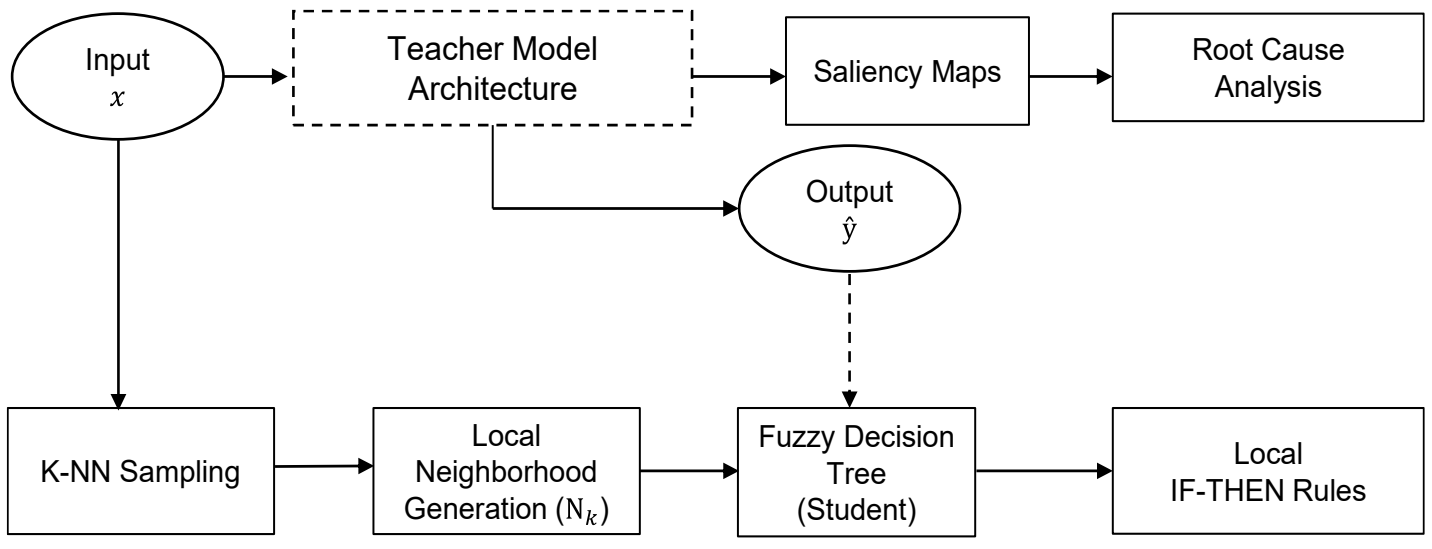

Figure 2: The testing phase of the proposed framework.

In particular, fuzzy sets are then defined in such a way to cover overlapping regions of the local neighborhood of each feature, using Gaussian membership functions, with set centers initialized via linear interpolation between the minimum and maximum feature values within the local neighborhood. In addition, the tree is constructed through recursive splitting guided by a weighted fuzzy entropy criterion [14], a process that encourages soft decision surfaces [15]. Each unique path from the root to a leaf node constitutes a fuzzy rule leading to class prediction $K$. In the proposed framework, final predictions are determined by aggregating the firing strength of the activated tree nodes, resulting in IF-THEN rules of the following form:

**IF** ($feature\ \boldsymbol{x_1}$) is *Low* **AND** ($feature\ \boldsymbol{x_2}$) is *High* **THEN** (prediction $\widehat{\boldsymbol{y}}$) belongs to (class $K$)

Each rule is characterized by a confidence value ($p$) Eq. (5) that quantifies the rule's local precision, defined as the proportion of the total samples for a prediction $\widehat{\boldsymbol{y}}$ that aligns with a predicted class $K$. This metric measures the rule's local precision, ensuring that the linguistic antecedents provide a robust and consistent mapping to the target label.

$$p(R_i) = \frac{\sum_{x \in \mathcal{D}} \alpha_{c_i}(x|\hat{y}=K)}{\sum_{x \in \mathcal{D}} \alpha_{c_i}(x)}, \ i = 1, \dots, \Phi \ (5)$$

where a rule $R_i$ corresponds to a root-to-leaf path and is associated with fuzzy regions in the feature space and $\Phi \in \mathbb{N}$ the total number of rules. Let $a_{c_i} \in [0,1]$ denote the firing strength (activation) of rule $R_i$ for input $\boldsymbol{x}$, *i.e.*, the aggregation of the membership degrees along the path, where $\boldsymbol{x} \in \mathcal{D}$ and $\mathcal{D}$ is the sample space.

# 3 EXPERIMENTS AND RESULTS

## 3.1 Dataset Description and Preprocessing

The dataset used in this study was generated from a highly sophisticated simulator, based on a Combined Diesel electric And Gas (CODLAG) propulsion plant mounted on a frigate. This simulator comprises various interconnected blocks, including Propeller, Hull, Gas Turbine (GT), Gear Box, and Controller, which have been meticulously developed and fine-tuned based on numerous real propulsion plants over time. The dataset reflects the behavior of the propulsion system, and it is characterized by two essential parameters: the compressor degradation coefficient $k_c$ and the turbine degradation coefficient $k_t$. Each degradation state is represented by a combination of these two coefficients. Specifically, the decay of GT ranges within $[0.975, 1.0]$, and includes 26 values, whereas the decay of Gas Turbine Compressor (GTC) ranges within $[0.95,1.0]$ and includes a total of 51 values. Additionally, the operational state is defined by the vessel speed, controlled by the lever position $l$. The lever assumes discrete values $l = 1, \dots, 9$ corresponding to ship speeds ranging from 3 to 27 knots in increments of 3 knots. Therefore, the space of all possible states is described by the following Cartesian product of the individual parameter sets:

$$\Omega = \{k_t\} \times \{k_c\} \times \{l\} \quad (6)$$

resulting in a total of $51 \times 26 \times 9 = 11934$ unique configuration states. Each configuration state yields a feature vector with $d = 18$ elements (including the 2 target degradation states $k_t$, $k_c$ (Eq. 7-8) and the lever position $l$), representing measurements from the vessel's continuous monitoring system.

$$k_c \begin{cases} [0.95, 0.98) & decayed \\ [0.98, 1.00] & not\ decayed \end{cases} \quad (7)$$

$$k_t \begin{cases} [0.97, 0.99) & decayed \\ [0.99, 1.00] & not\ decayed \end{cases} \quad (8)$$

This vector characterizes the propulsion system's behavior under varying operational and health conditions. To facilitate a condition-based PdM framework, we adopt a binary classification approach by thresholding the target degradation states, following the methodology proposed by Cipollini et al. [4]. To express the health status of the GT $k_t$ and GTC $k_c$ the following discretization is performed into the space $y \in \{0,1\}$, as proposed in [4]:

$$y_{k_c} = \begin{cases} 1, & 0.95 \le k_c < 0.98 \\ 0, & 0.98 \le k_c \le 1.00 \end{cases} \quad (9)$$

$$y_{k_t} = \begin{cases} 1, & 0.97 \le k_t < 0.99 \\ 0, & 0.99 \le k_t \le 1.00 \end{cases} \quad (10)$$

## 3.2 Classification Results

In order to evaluate the performance of the proposed model the following metrics were used: accuracy, sensitivity, specificity, and Area under the Receiving Operating Characteristic (AUC) [16]. The performance of the proposed framework was compared to several baseline approaches, including Logistic Regression [17], Support Vector Machines (SVM) [18], the Classification and Regression Tree (CART) algorithm [19], Random Forest [20], the CatBoost classifier [21], as well as the state-of-the-art Modern Neighborhood Components Analysis (M-NCA) [22]. For logistic regression the regularization parameter $C$ was tuned over $[10^{-4}, 10^{2}]$, whereas the SVM was evaluated using both linear and Radial Basis Function (RBF) kernels, with $C \in [10^{-3}, 10^{2}]$, and for the RBF kernel $\gamma \in [10^{-4}, 10]$. For tree-based models, the maximum tree depth was set within [5,20] for both CART and RF; the number of estimators for RF was tuned over the interval [100,500]. For CatBoost, the tree depth, border count, and bagging temperature were optimized within the intervals [4,14], [32,255], and $[10^{-4}$,10], respectively. NN-based models, including M-NCA and the proposed one, were optimized over the number of layers, hidden layer dimensions, projection dimensions, and embedding dimensions, within the intervals [1,4], [64,256], [32,128], and [128,256], respectively.

The proposed framework was evaluated using a 5-fold stratified cross-validation. The results for the compressor and the turbine on the testing set are presented in Tables 2 and 3, respectively. As can be observed from the tables, the proposed framework provides better results compared to the other methods, except for SVM and M-NCA which have comparable results. Specifically, for the case of the compressor, the proposed framework achieved an accuracy of 99.24%, with a 99.43% sensitivity and a specificity equal to 98.90%. Furthermore, the AUC was 99.20%.

Table 2: Comparisons of the proposed framework in terms of Accuracy (%), AUC (%), Specificity (%), and Sensitivity (%) with well-known and state-of-the-art classifiers for the compressor.

| Models/Metrics | Accuracy (%) | AUC (%) | Specificity (%) | Sensitivity (%) |
|---|---|---|---|---|
| Logistic Regression [17] | 93.52 % | 93.43 % | 92.35 % | 94.52 % |
| SVM [18] | 99.37 % | 99.20 % | 99.18 % | 99.21 % |
| CART [19] | 96.81 % | 96.72 % | 96.22 % | 97.22 % |
| Random Forest [20] | 98.20 % | 98.09 % | 97.24 % | 98.90 % |
| CatBoost [21] | 93.79 % | 93.21 % | 89.90 % | 96.51 % |
| M-NCA [22] | 99.74 % | 99.72 % | 99.59 % | 99.85 % |
| **Proposed framework** | **99.24 %** | **99.20 %** | **98.90 %** | **99.43 %** |

Table 3: Comparisons of the proposed framework in terms of Accuracy, AUC, Specificity, and Sensitivity, with well-known and state-of-the-art classifiers for the turbine.

| Models/Metrics | Accuracy (%) | AUC (%) | Specificity (%) | Sensitivity (%) |
|---|---|---|---|---|
| Logistic Regression[17] | 95.18 % | 95.03 % | 94.05 % | 96.00 % |
| SVM [18] | 99.30 % | 99.17 % | 99.00 % | 99.20 % |
| CART [19] | 91.78 % | 91.32 % | 91.00 % | 94.33 % |
| Random Forest [20] | 98.15 % | 98.00 % | 97.02 % | 98.98 % |
| CatBoost [21] | 93.75 % | 93.37 % | 90.89 % | 95.86 % |
| M-NCA [22] | 99.58 % | 99.57 % | 99.50 % | 99.63 % |
| **Proposed framework** | **98.91 %** | **99.05 %** | **99.78 %** | **98.11 %** |

Regarding the turbine, the proposed framework achieved accuracy equal to 98.91%, 99.78% specificity, 98.11% sensitivity, and an AUC value of 99.05%. Furthermore, it has to be mentioned that an advantage over the rest compared approaches is that the proposed framework provides outcomes that are understandable in a way that is compatible with human perception.

### 3.3 Example of Explainable Failure Prediction

The proposed framework provides an explainable failure prediction, while discovering relations among the examined coefficients and analyzing them, using cause-and-effect relations. To better understand the proposed framework, two indicative examples are presented, corresponding to a healthy and a degraded compressor state under a lever position configuration of 3. In this context, the variable *GGn* represents the rate of revolutions (measured in rpm) of the gas generator, and *T2* corresponds to the outlet air temperature (measured in C) of the gas turbine compressor. Following the methodology described in Section 2.1, the teacher model initially infers each test sample and generates soft labels. A local neighborhood $\boldsymbol{N_k}$ is constructed and used, together with the soft labels, to train the FDT (Section 2.2). The FDT leverages local membership functions to provide interpretable explanations and employs a fuzzy information gain criterion to reduce the feature space to the most influential variables. The membership functions presented in Figure 3 are obtained.

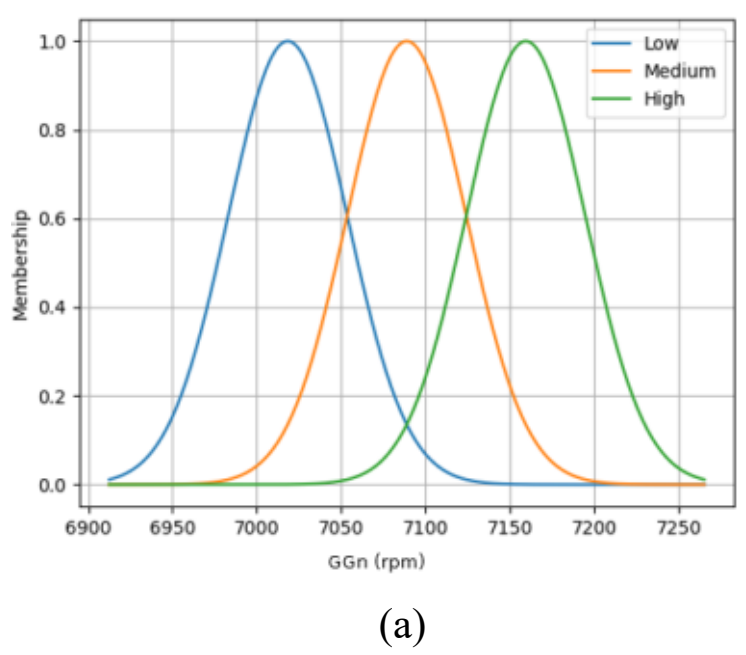

(a)

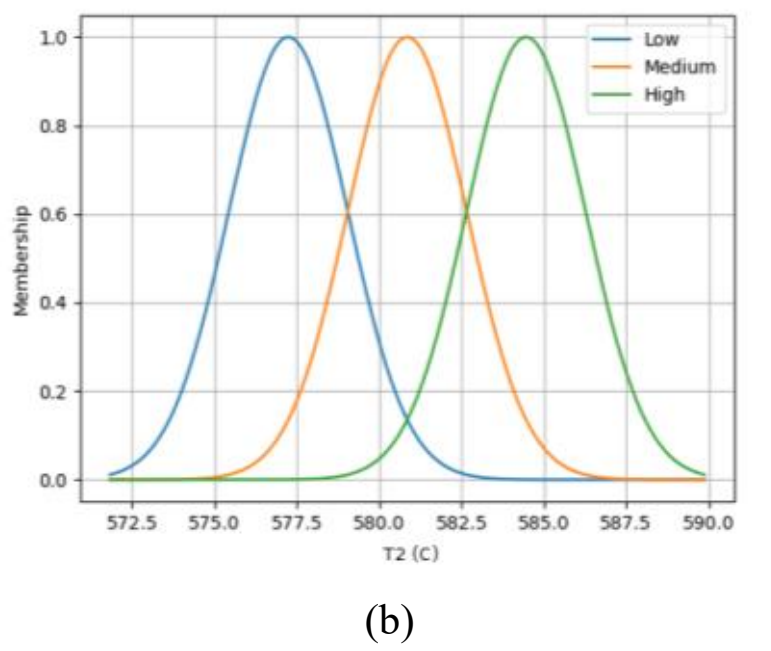

(b)

Figure 3: Membership functions for different variables. (a) GGn. (b) T2.

The trained FDT extracts fuzzy rules that characterize the local decision behavior within $\boldsymbol{N_k}$. Each rule corresponds to a leaf node of the tree and provides a localized explanation of the model's prediction $\hat{\boldsymbol{y}}$. From the extracted rule set, only the rules associated with the predicted class of interest are retained. In the following examples, two FDTs are utilized; the confidence value ($p$) of each rule, reported in parentheses, reflects its predictive reliability. The first sample is related to a Healthy Compressor State (HCS) characterized by $GGn = 7068.72rpm$ and $T2 = 577.579°C$, whereas the second represents a Degraded Compressor State (DCS) characterized by $GGn = 7122.785rpm$, $T2 = 582.707°C$. The first FDT extracts 5 fuzzy rules regarding the Healthy Compressor State, and the second FDT generates 6 fuzzy rules associated with the Degraded Compressor State. From the total 11 extracted rules, the most informative rules for each state, *i.e.*, Healthy Compressor State: Rules (HCS) 1-3, Degraded Compressor State: Rules (DCS) 4-6, are presented below:

Rule 1 (HCS): **IF** *GGn* is *Low* **THEN** the Compressor State is *Healthy* ($p = 0.94$)

Rule 2 (HCS)a: **IF** *T2* is *Low* **AND** *GGn* is *Medium* **THEN** the Compressor State is *Healthy* ($p = 0.77$)

Rule 3 (HCS): **IF** *T2* is *Low* **AND** *GGn* is *Medium* **THEN** the Compressor State is *Healthy* ($p = 0.56$)

Rule 1 (DCS): **IF** *T2* is *High* **THEN** the Compressor State is *Degraded* ($p = 0.61$)

Rule 2 (DCS): **IF** *T2* is (*Medium* **OR** *High*) AND *GGn* is *Medium* **THEN** the Compressor State is *Degraded* ($p = 0.82$)

Rule 3 (DCS): **IF** *T2* is *Medium* **AND** *GGn* is *High* **THEN** the Compressor State is *Degraded* ($p = 0.97$)

The extracted rules indicate that *Low* values of *T2* and *GGn* are strongly associated with healthy compressor behavior, whereas degraded state predictions are mainly associated with *High* values of *T2,* which can also be combined with *Medium* or *High* values of *GGn*.

## 4 CONCLUSIONS

In this paper a novel framework based on a fuzzy decision tree and a deep residual neural network was proposed, aiming to perform explainable PdM on propulsion systems of naval vessels. The proposed framework is able to generate fuzzy local rules based on the dataset used. Furthermore, it can provide explainable predictions by integrating differentiable fuzzy layers into the core predictive model and complementing them with local instance-specific fuzzy rules. In this way, the proposed framework can effectively provide understandable explanations using cause-and-effect relationships to the users, thus gaining their trust. The experiments demonstrated the effectiveness of the proposed framework over other state-of-the-art and baseline approaches validating its capabilities on PdM tasks for NPSs. Notably, even when predictive performance is on par with other high-performing approaches, the proposed framework offers a significant advantage through its inherent explainability. For PdM tasks, the framework provides explainable results through IF-THEN rules, in conjunction with a component's health state predictions, providing human-readable interpretations of the outcomes. As a result, the methodology not only maintains competitive predictive accuracy but also enhances decision-making by revealing the underlying causes of component degradation, enabling engineers to more effectively identify issues and target necessary repairs. Future research includes the application of the proposed framework for PdM of other sea vessel components, as well as its adaptation and application on other domains, where explainable predictive modeling can also be critical for proactive decisions, such as aviation and healthcare.

## ACKNOWLEDGMENTS

This work has received funding by the European Union (EU)'s Horizon Europe research and innovation programme under grant agreement No101202933 (D-NAVIO). Views and opinions expressed are however those of the authors only and do not necessarily reflect those of the EU or the European Climate, Infrastructure, and Environment Executive Agency (CINEA). Neither the European Union nor the granting authority can be held responsible for them.

## REFERENCES

[1] F. Javadnejad, H. J. Park, S. Kovacic, and A. Sousa-Poza, "Predictive maintenance in naval vessel propulsion systems for enhanced marine operations using a BiGMM-HMM framework with divergence-based clustering," *Discover Oceans*, vol. 2, no. 1, p. 33, 2025.

[2] D. D. M. Frangopol, D. P. Bocchini, A. Decò, D. S. Kim, D. K. Kwon, D. N. M. Okasha, and D. Saydam, "Integrated life-cycle framework for maintenance, monitoring, and reliability of naval ship structures," *Naval Engineers Journal*, vol. 124, no. 1, pp. 89–99, 2012.

[3] K. Kwon, D. M. Frangopol, and S. Kim, "Fatigue performance assessment and service life prediction of high-speed ship structures based on probabilistic lifetime sea loads," *Structure and Infrastructure Engineering*, vol. 9, no. 2, pp. 102–115, 2013.

[4] F. Cipollini, L. Oneto, A. Coraddu, A. J. Murphy, and D. Anguita, "Condition-based maintenance of naval propulsion systems: Data analysis with minimal feedback," *Reliability Engineering & System Safety*, vol. 177, pp. 12–23, 2018.

[5] J. Gao, S. Dong, J. Cui, M. Yuan, and J. Zhao, "DRN-GAN: an integrated deep learning-based health degradation assessment model for naval propulsion system," *Engineering Computations*, vol. 39, no. 6, pp. 2306–2325, 2022.

[6] F. Javadnejad, H. J. Park, S. Kovacic, and A. Sousa-Poza, "A Novel BiGMM-HMM Framework for Predictive Maintenance in Naval Vessel Propulsion Equipment," in *2024 19th Annual System of Systems Engineering Conference (SoSE)*, 2024, pp. 116–123.

[7] M. A. Sahraoui, C. Rahmoune, A. Damou, F. Gougam, and A. Afia, "Advancing condition-based maintenance of naval propulsion systems with ensemble learning techniques," *Advances in Mechanical Engineering*, vol. 16, no. 11, p. 16878132241298373, 2024.

[8] K. He, X. Zhang, S. Ren, and J. Sun, "Deep residual learning for image recognition," in *Proceedings of the IEEE conference on computer vision and pattern recognition*, 2016, pp. 770–778.

[9] J.-S. Jang, "ANFIS: adaptive-network-based fuzzy inference system," *IEEE transactions on systems, man, and cybernetics*, vol. 23, no. 3, pp. 665–685, 1993.

[10] R. Das, S. Sen, and U. Maulik, "A survey on fuzzy deep neural networks," *ACM Computing Surveys (CSUR)*, vol. 53, no. 3, pp. 1–25, 2020.

[11] M. Yeganejou, R. Kluzinski, S. Dick, and J. Miller, "An end-to-end trainable deep convolutional neuro-fuzzy classifier," in *2022 IEEE International Conference on Fuzzy Systems (FUZZ-IEEE)*, 2022, pp. 1–7.

[12] M. Yeganejou, K. Honari, R. Kluzinski, S. Dick, M. Lipsett, and J. Miller, "Dcnfis: Deep convolutional neuro-fuzzy inference system," *arXiv preprint arXiv:2308.06378*, 2023.

[13] K. Simonyan, A. Vedaldi, and A. Zisserman, "Deep inside convolutional networks: Visualising image classification models and saliency maps," *arXiv preprint arXiv:1312.6034*, 2013.

[14] C. Z. Janikow, "Fuzzy decision trees: issues and methods," *IEEE Transactions on Systems, Man, and Cybernetics, Part B (Cybernetics)*, vol. 28, no. 1, pp. 1–14, 1998.

[15] C. Olaru and L. Wehenkel, "A complete fuzzy decision tree technique," *Fuzzy sets and systems*, vol. 138, no. 2, pp. 221–254, 2003.

[16] T. Fawcett, "An introduction to ROC analysis," *Pattern recognition letters*, vol. 27, no. 8, pp. 861–874, 2006.

[17] F. Pedregosa, G. Varoquaux, A. Gramfort, V. Michel, B. Thirion, O. Grisel, M. Blondel, P. Prettenhofer, R. Weiss, V. Dubourg, and others, "Scikit-learn: Machine learning in Python," *the Journal of machine Learning research*, vol. 12, pp. 2825–2830, 2011.

[18] B. E. Boser, I. M. Guyon, and V. N. Vapnik, "A training algorithm for optimal margin classifiers," in *Proceedings of the fifth annual workshop on Computational learning theory*, 1992, pp. 144–152.

[19] T. Hastie, R. Tibshirani, and J. Friedman, "An introduction to statistical learning," 2009.

[20] L. Breiman, "Random forests," *Machine learning*, vol. 45, no. 1, pp. 5–32, 2001.

[21] L. Prokhorenkova, G. Gusev, A. Vorobev, A. V. Dorogush, and A. Gulin, "CatBoost: unbiased boosting with categorical features," *Advances in neural information processing systems*, vol. 31, 2018.

[22] H.-J. Ye, H.-H. Yin, D.-C. Zhan, and W.-L. Chao, "Revisiting nearest neighbor for tabular data: A deep tabular baseline two decades later," *arXiv preprint arXiv:2407.03257*, 2024.